\documentclass[letterpaper]{article} 
\newif\ifanon
\anonfalse
\ifanon
\usepackage[submission]{aaai25}  
\else
\usepackage{aaai25}
\nocopyright
\fi
\usepackage{times}  
\usepackage{helvet}  
\usepackage{courier}  
\usepackage[hyphens]{url}  
\usepackage{graphicx} 
\usepackage{natbib}  
\usepackage{caption} 
\usepackage{amsmath}
\usepackage{amssymb}
\usepackage{booktabs}
\usepackage{xcolor}
\newcommand{\answerYes}[1]{\textcolor{blue}{#1}}
\newcommand{\answerNo}[1]{\textcolor{teal}{#1}}
\newcommand{\answerNA}[1]{\textcolor{gray}{#1}}
\title{Embedding Cosine Measures Text Length Before It Measures Homogenization}
\ifanon
\author{Anonymous Submission}
\affiliations{}
\else
\author{Fengming Liu\textsuperscript{\rm 1}, Mingyu Shu\textsuperscript{\rm 2}}
\affiliations{\textsuperscript{\rm 1}University of Chicago\\
\textsuperscript{\rm 2}School of Business, Anhui University\\
fengmingliu@uchicago.edu, shumingyu@foxmail.com}
\fi

\begin{document}
\maketitle

\begin{abstract}
Mean pairwise cosine similarity between transformer embeddings is a standard measure of whether online discourse is converging under generative AI. When the embeddings are mean-pooled outputs of an encoder never trained for sentence similarity, this measure mostly tracks text length. In 6.06 million Reddit comments, an uncentered DeBERTa-v3 measure had suggested that only conservative discourse broke toward homogeneity after late 2022. Truncating the same comments from 10 to 150 words raises their mean cosine by eight times the reported break. A placebo that swaps every comment for a random one of equal length reproduces 81\% of the asymmetry, and a fixed-content simulation reproduces 78\%. The leading principal direction carries 99.9\% of raw DeBERTa homogeneity, and once length is handled the asymmetry disappears under every encoder. On 575{,}000 Stack Exchange answers the measure reports a post-ChatGPT homogenization of technical sites that the placebo fully reproduces, although answers grew longer. A census of 31 studies locates the risk where anisotropic encoders meet time-series or length-unequal designs. A reverse control on a well-known experiment shows that centering with renormalization can reverse a genuine effect. We propose a low-cost diagnostic protocol. Corrected measures show that the two political camps converged between 2019 and 2022 and stopped converging before ChatGPT. Cross-camp replies engage their targets more closely than same-camp replies.
\end{abstract}

\section{Introduction}

Studies of AI-assisted writing measure homogeneity as the mean cosine between document embeddings, and with it they show that model help narrows the range of stories and ideas a group produces \citep{doshi2024,padmakumar2024,anderson2024,agarwal2025}. The same construction now travels from controlled experiments to observational platform data, where it asks whether whole sites and forums have converged since ChatGPT \citep{sourati2025,delriochanona2024}. It gives one number per forum per month for whether people are coming to say the same things, which hand coding cannot deliver. It is cheap, scales to millions of documents, and appears to capture meaning beyond vocabulary.

That number is a cosine, and the cosine between two embeddings is a property of the embedding space before it is a property of the texts. Contextual encoders produce anisotropic representations that crowd into a narrow cone, so two unrelated sentences already share a high cosine \citep{ethayarajh2019,gao2019,li2020bertflow}. A handful of rogue dimensions can dominate the similarity \citep{timkey2021}, cosine can encode token frequency in place of meaning \citep{zhou2022cosine}, and its values depend on training choices unrelated to semantic closeness \citep{steck2024cosine}. Embeddings of longer texts also collapse toward one another irrespective of what they say \citep{zhou2025lengthcollapse,xiao2023length}. The representation-learning literature has established all of this. What we document here is how completely these properties can determine a substantive social-science finding once the measure enters a time series.

The case is an analysis of r/AskConservatives and r/AskALiberal from 2019 to 2025 \citep{liu2026asym}\ifanon\else, the first author's earlier study.\fi{}
Its headline result was an asymmetric break. Within-group similarity among right-flaired users in r/AskConservatives rose above its pre-existing downward trend after December 2022 ($\gamma=+0.0081$, $p<0.001$), while left-flaired users in r/AskALiberal moved far less. The measure was the mean pairwise cosine of mean-pooled DeBERTa-v3-base embeddings \citep{he2023debertav3}. We re-estimate that result on the same corpus with six diagnostics and two sentence encoders and find a length artifact. Right-flaired comments in r/AskConservatives shrank from 109 words on average in 2019 to 61 in 2022 and then stabilized. Under an anisotropic measure the falling length line looks like a diversification trend, and its flattening looks like a break.

Stack Exchange, a second platform, shows the same spurious break with the length shift running in the opposite direction. An exact spectral decomposition explains both cases, because the pairwise mean is a sum over principal directions and the length signal sits in the leading one. A census of 31 published and preprinted studies maps where else the risk lies, and a reverse control on the best-known experiment in this literature \citep{doshi2024} shows that the usual remedy carries its own danger. Any study that reads change in raw embedding cosine as change in discourse is exposed whenever the length distribution of its texts moves, and on social platforms that distribution nearly always moves. We show how large the artifact is, how to detect it with a placebo that costs nothing beyond the embeddings already computed, and which corrections remove it without creating new artifacts.
With the measure repaired, the corpus shows convergence between the two camps before ChatGPT and closer engagement across the divide than within it.

\section{Related Work}

\paragraph{Embedding similarity as a measure of homogenization.} Writers who receive suggestions from a language model produce stories or essays that are more similar to one another than those written without help, with similarity measured as the average cosine between document embeddings \citep{doshi2024,padmakumar2024}. Related experiments find the same narrowing in ideation and creative tasks \citep{anderson2024} and across cultures \citep{agarwal2025}. The concern has since been framed as a societal risk to the diversity of expression and thought \citep{sourati2025}, and observational studies now ask whether platform activity has changed since ChatGPT \citep{delriochanona2024}. In observational designs the comparison runs across time, and everything else that changes over time, including how long people write, enters the measure.

\paragraph{Geometry of contextual embeddings.} Contextual embeddings occupy a narrow cone in which unrelated inputs share large cosines \citep{ethayarajh2019,gao2019}. Proposed remedies for this anisotropy include removing the dominant directions \citep{mu2018allbutthetop}, whitening \citep{su2021whitening}, flow-based normalization \citep{li2020bertflow}, and cluster-based corrections \citep{rajaee2021cluster}, and isotropy can be measured directly \citep{rudman2022isoscore}. A few rogue dimensions can dominate cosine similarity \citep{timkey2021}, cosine is distorted by word frequency \citep{zhou2022cosine}, and its meaning depends on regularization choices made during training \citep{steck2024cosine}. Two recent papers show that embedding similarity also depends on text length, with longer inputs collapsing toward one another \citep{xiao2023length,zhou2025lengthcollapse}. Sentence encoders trained with contrastive objectives were built to address the failure of mean-pooled masked language models on similarity tasks \citep{reimers2019sbert,xiao2024cpack}.

\paragraph{Validating text measures.} Computational social science has long insisted that automated text measures be validated against the construct they claim to capture \citep{grimmer2013,rodriguez2022}, and measurement modeling offers a vocabulary for separating construct validity from reliability \citep{jacobs2021measurement}. Diversity measurement has its own standardization effort \citep{shaib2025diversity}, and eigenvalue-based scores such as the Vendi score offer alternatives to the pairwise mean \citep{friedman2023vendi}. Studies of AI influence on text have also used distributional and lexical markers that do not depend on embedding geometry \citep{liang2024monitoring,liang2025widespread,kobak2025}. We take the known geometric properties of embeddings and show, in a full applied case, how they generate a substantive time-series finding and how a study can detect that before publication.

\section{Data and Measures}

The corpus contains 6{,}056{,}593 comments posted to r/AskConservatives and r/AskALiberal between January 2019 and December 2025, with ideology assigned from user flair into Left and Right. The four subreddit-by-ideology groups are large and unequal. Left-flaired users in r/AskALiberal contributed 2.98 million comments and right-flaired users in r/AskConservatives 1.66 million. For every subreddit $\times$ ideology $\times$ month cell we draw a stratified random sample of up to 3{,}000 comments (seed fixed), yielding 915{,}225 comments across all 336 cells. Every diagnostic below runs on this one sample, so differences between measures cannot come from differences in sampling. To rule out the sample itself as a source of error, we also encode all 6{,}056{,}593 comments with mpnet and compute the raw, centered, and length-stratified measures exactly from per-cell vector sums.

We encode the sample three times. The first encoding reproduces the original pipeline, mean-pooling the last hidden layer of DeBERTa-v3-base over at most 256 tokens. The second and third use encoders trained for sentence similarity, all-mpnet-base-v2 \citep{reimers2019sbert,song2020mpnet} and BGE-base-en-v1.5 \citep{xiao2024cpack}. All vectors are L2-normalized.

For a cell with normalized embeddings $x_1,\dots,x_n$ and sum vector $s=\sum_i x_i$, the raw homogeneity measure is
\begin{equation}
H=\frac{1}{n(n-1)}\sum_{i\neq j}x_i^\top x_j=\frac{s^\top s-n}{n(n-1)},
\end{equation}
which makes plain that $H$ is, up to $O(1/n)$, the squared length of the cell's mean vector. Any shared component of the embeddings inflates it. For unit vectors $x_i^\top x_j=1-\tfrac12\lVert x_i-x_j\rVert^2$, so $H$ is also one minus half the mean squared distance between comments, and subtracting any common vector without renormalizing leaves it unchanged. The corrected measure that carries every substantive result below keeps this form and handles length by stratification. It computes $H$ on the uncentered vectors separately within global word-count quintiles (cut at 15, 28, 47, and 86 words), requires at least five comments in every quintile, and averages the five values with equal weight. As diagnostic contrasts we compute four further variants and the Vendi score. The centered measure subtracts the corpus mean vector before renormalizing, the simplest correction for anisotropy \citep{mu2018allbutthetop,su2021whitening}, and we show later that its renormalization step can reverse a genuine effect. The centered length-tercile measure computes the centered measure separately within word-count terciles and averages them. The cross-thread measure keeps only pairs of comments from different submissions, which removes similarity produced by shared topics. The author-balanced measure collapses each author's comments in a month to one mean vector, so that a few prolific posters cannot dominate. The Vendi score \citep{friedman2023vendi} is an eigenvalue-based alternative, and it behaves like the centered measure throughout.

For the asymmetry we estimate the original difference-in-differences on the two own-ideology groups (right-flaired in r/AskConservatives, left-flaired in r/AskALiberal), with group-specific linear trends, a post-December-2022 indicator, and their interaction, using Newey--West standard errors with six lags. We express the outcome in units of each series' monthly standard deviation in 2021--22 so that measures on different scales can be compared. Because the choice of pre-period turns out to matter, we report every estimate twice, with a baseline starting in January 2019 and with one starting in January 2021 \citep{hausman2018rdit,bernal2017its}.

\section{Diagnostic 1: The Same Text at Different Lengths}

We draw 20{,}000 comments of at least 150 words, truncate each to its first $k$ words for $k\in\{10,20,40,80,150\}$, and compute $H$ for each truncation level. The comments, authors, and topics are identical across levels, so only length varies.

\begin{figure}[t]
\centering
\includegraphics[width=\columnwidth]{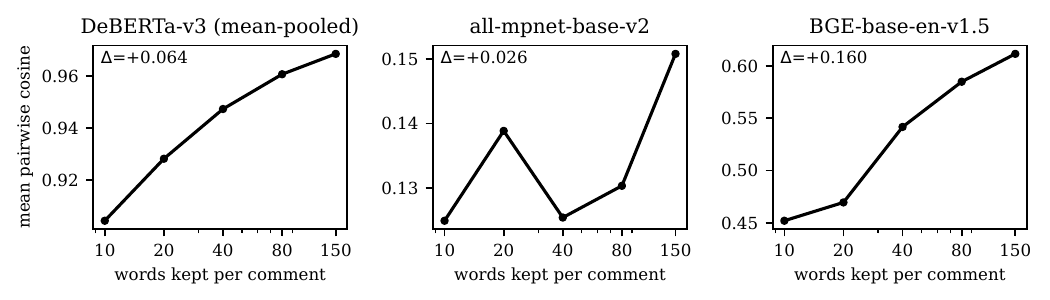}
\caption{Mean pairwise cosine of the same 20{,}000 comments truncated to $k$ words. Content is held fixed and only length changes.}
\label{fig:trunc}
\end{figure}

Under DeBERTa, $H$ rises monotonically from 0.904 at ten words to 0.968 at 150 words (Figure~\ref{fig:trunc}). The increase of 0.064 is eight times the reported conservative break of 0.0081. BGE shows a larger absolute increase, from 0.452 to 0.612, so a sentence encoder does not by itself protect against length. Only mpnet stays nearly flat, moving between 0.125 and 0.151 without a monotone pattern. The same ordering appears among naturally occurring comments. Random pairs drawn from the same length bin average 0.879 under DeBERTa when both comments have fewer than 15 words and 0.970 when both have more than 120.

Random pairs of DeBERTa embeddings in the corpus share a mean cosine of 0.924, against 0.111 for mpnet and 0.520 for BGE. The first principal component of the DeBERTa embeddings carries 32\% of total variance, and the first ten carry 66\%, against 3--4\% and 21--23\% for the two sentence encoders. A comment's cosine with the corpus mean direction correlates with its log length at 0.52 under DeBERTa and 0.60 under BGE. Longer comments sit closer to the common direction, so a measure that is the squared length of a mean vector rises whenever comments get longer.

After subtracting the corpus mean, DeBERTa similarity at fixed truncation is U-shaped in length (0.337 at ten words, 0.026 at forty, 0.321 at 150), because very short and very long comments each occupy their own region away from the global mean. Centering leaves a length dependence of its own.

\section{Diagnostic 2: A Length-Only Placebo}

For every cell we replace each sampled comment with a randomly drawn comment from the same half-octave word-count bin, taken from the pooled sample of all groups and months. The placebo cell keeps its length distribution and loses everything else, including topic, ideology, period, and author. We recompute $H$ on the placebo cells, rerun the full inference, and repeat the procedure fifty times, which gives the share of the published series that length alone explains.

\begin{figure}[t]
\centering
\includegraphics[width=\columnwidth]{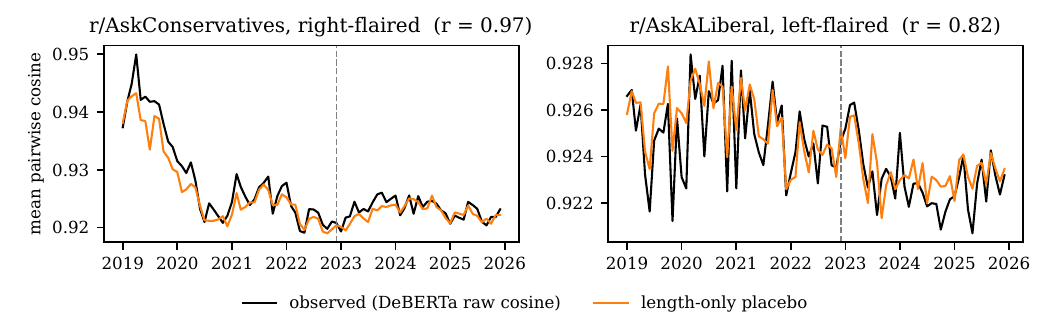}
\caption{Observed DeBERTa raw homogeneity (black) and the mean of fifty length-only placebo series (orange). Dashed line marks December 2022.}
\label{fig:placebo}
\end{figure}

For right-flaired users in r/AskConservatives the placebo series correlates with the observed series at $r=0.97$ (Figure~\ref{fig:placebo}). It reproduces the steep decline of 2019--20, the floor of 2022, and the flat post-period that the original design read as a departure from trend. For left-flaired users in r/AskALiberal the correlation is 0.82. The observed series also correlates with the cell's mean word count at 0.95 for the conservative group. Even the crudest control, adding mean word count and its interaction with group to the difference-in-differences, cuts the asymmetry from 0.0089 ($p=0.021$) to 0.0024 ($p=0.073$).

In raw units, the Right-versus-Left difference-in-differences on the DeBERTa sample is 0.0089 ($p=0.021$). The placebo yields 0.0072 on average across draws (SD 0.0007), and 58\% of placebo draws reach $p<0.05$. Length alone therefore reproduces 81\% of the asymmetry. The conservative ITS level shift tells the same story, 0.0062 observed against 0.0051 from the placebo. When we regress the observed series on its placebo and rerun the analysis on the residual, the asymmetry falls to 0.59 SD ($p=0.48$) under DeBERTa, $-0.58$ ($p=0.65$) under mpnet, and $-0.50$ ($p=0.66$) under BGE.

\section{Diagnostic 3: A Fixed-Content Simulation}

Placebo donors are still real comments, and a short comment in 2019 may differ in content from a short comment in 2024. The simulation fixes content entirely. We draw a pool of 8{,}000 comments of at least 200 words and encode each one at thirteen truncation lengths between 3 and 200 words. A synthetic cell for any group and month then draws 1{,}000 comments from this one pool and truncates each to a length drawn from that cell's observed word-count distribution. Every synthetic cell has the same topics, the same authors, and the same ideological mix in expectation. Only the distribution of lengths differs. We repeat the construction twenty times.

The first scenario feeds in the observed length distributions. The fixed-content series correlates with the observed right-flaired series in r/AskConservatives at $r=0.97$ under DeBERTa, $0.95$ under BGE, and $0.67$ under mpnet. Run through the same difference-in-differences, it produces an asymmetry of 0.0069 under DeBERTa against the observed 0.0089 ($p=0.04$ for the averaged series), so content-free length variation reproduces 78\% of the effect. Under BGE the share is 96\% (0.0144 against 0.0151), and under mpnet the simulated and observed estimates are equal (0.0032 and 0.0031). Even the point estimates of the centered measures, which were not significant in the real data, are 66--85\% length. Most of any residual asymmetry in a corrected measure should therefore be attributed to length unless length is controlled directly.

\begin{figure*}[t]
\centering
\includegraphics[width=0.85\textwidth]{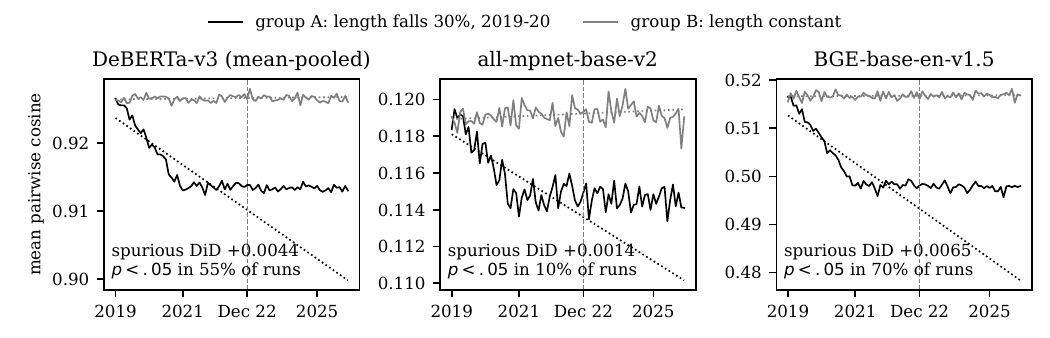}
\caption{Fixed-content simulation. Two groups draw comments from one pool. Group A's comment length falls by 30\% across 2019--20 and then stays flat, group B's stays constant. Dotted lines are linear trends fitted on the pre-December-2022 period. Mean of twenty runs.}
\label{fig:sim}
\end{figure*}

The second scenario is stylized, so that the mechanism can be seen in isolation (Figure~\ref{fig:sim}). Two groups draw from the same pool. Group A's median length falls by 30\% over the first two years and then stays flat, and group B's length never changes. Nothing happens in December 2022. The raw measure nonetheless produces a positive difference-in-differences at that date in every model, significant at $p<0.05$ in 55\% of runs under DeBERTa, 70\% under BGE, and 10\% under mpnet. In the figure, a linear trend fitted to the kinked series extrapolates the early decline into a period where the decline has stopped, and the gap between the extrapolation and the flat series is read as a level shift. Centering removes the spurious positive effect for DeBERTa (5\% of runs significant), but under the sentence encoders it produces a small spurious negative effect, significant in 45--70\% of runs because the simulated series are very smooth.

\begin{figure}[t]
\centering
\includegraphics[width=\columnwidth]{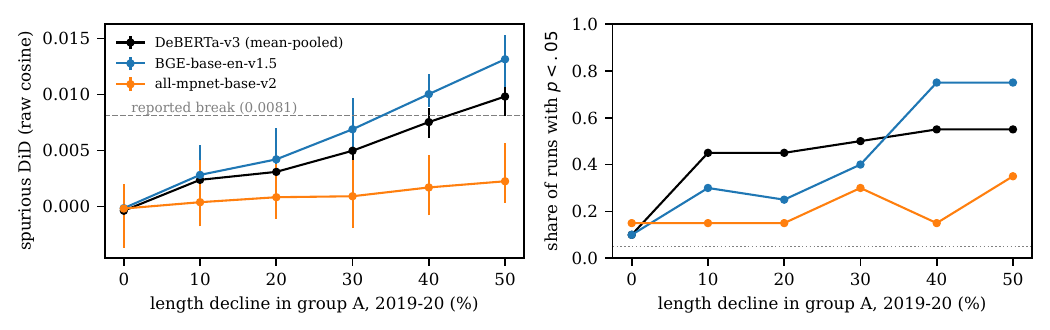}
\caption{Calibration of the stylized scenario. Left, spurious difference-in-differences on the raw measure as a function of the early length decline in group A (mean and 5th--95th percentile over twenty runs). Right, share of runs rejecting at $p<.05$.}
\label{fig:calib}
\end{figure}

Varying the size of the early decline turns this into a calibration curve (Figure~\ref{fig:calib}). Under DeBERTa, each ten-point decline in group A's comment length adds roughly 0.002 to the spurious estimate, and a decline of 40--50\% alone matches the reported break of 0.0081. The observed decline among right-flaired users in r/AskConservatives was about 29\% between the 2019--20 and 2021--22 averages (94 to 67 words), which the curve maps to a spurious estimate of about 0.005, more than half of the observed asymmetry before any other feature of the length distribution is considered. BGE is more sensitive still, and mpnet's raw measure is the least affected. With no length change at all, 10--15\% of runs reject at $p<0.05$, two to three times the nominal rate, because 84 monthly observations with six Newey--West lags give anti-conservative intervals for a level shift. A $p$-value near 0.02 from this design is weak evidence even with a sound measure.

\section{Diagnostic 4: The Asymmetry Across Measures}

\begin{figure*}[t]
\centering
\includegraphics[width=0.8\textwidth]{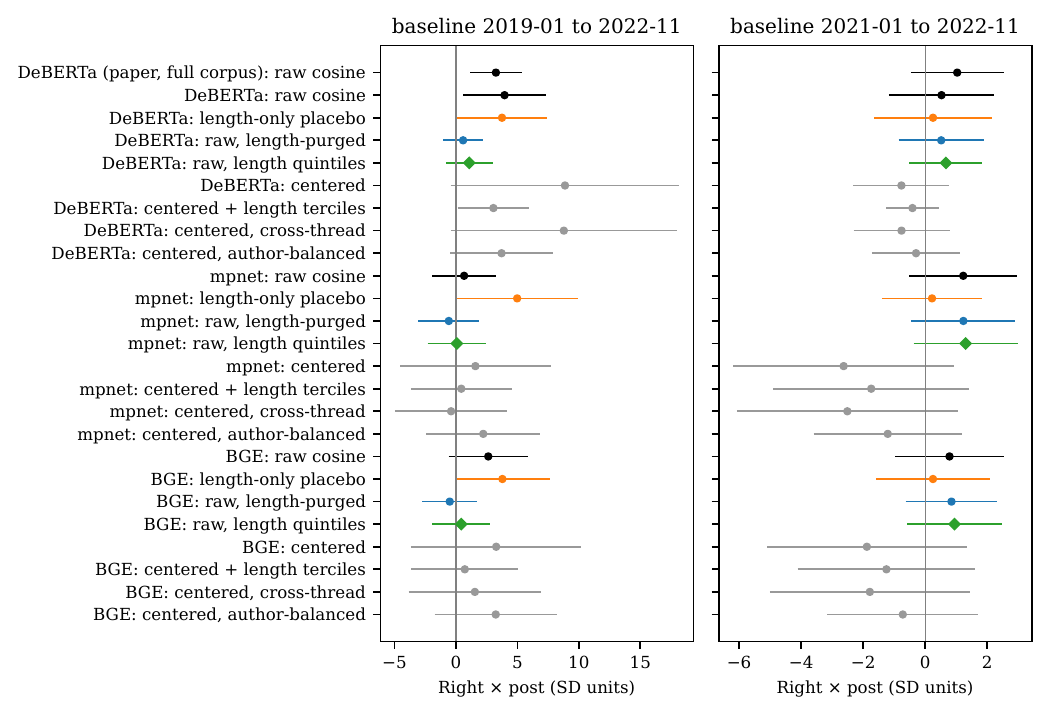}
\caption{Right $\times$ post difference-in-differences with 95\% Newey--West intervals, in SD units of the 2021--22 series. Black is raw cosine, orange the length-only placebo, blue the placebo-partialled series, green diamonds the corrected measure, and grey the centered and renormalized variants.}
\label{fig:forest}
\end{figure*}

\begin{table*}[t]
\centering\small
\begin{tabular}{lrrrrcrrrr}
\toprule
 & \multicolumn{4}{c}{Baseline 2019-01} & & \multicolumn{4}{c}{Baseline 2021-01} \\
\cmidrule{2-5} \cmidrule{7-10}
Measure & DeBERTa & mpnet & BGE & mpnet (all) & & DeBERTa & mpnet & BGE & mpnet (all) \\
\midrule
Original (full corpus) & +3.26$^{**}$ &  &  &  &  & +1.03 &  &  &  \\
Raw cosine & +3.96$^{*}$ & +0.67 & +2.63 & +0.63 & & +0.53 & +1.23 & +0.78 & +1.21 \\
Length-only placebo & +3.75$^{*}$ & +4.98$^{*}$ & +3.78$^{\dagger}$ & -- & & +0.25 & +0.22 & +0.25 & -- \\
Raw, placebo partialled out & +0.59 & $-$0.58 & $-$0.50 & -- & & +0.52 & +1.23 & +0.85 & -- \\
Raw, length quintiles (corrected) & +1.09 & +0.08 & +0.44 & +0.00 & & +0.67 & +1.31 & +0.95 & +1.31 \\
\addlinespace[2pt]\multicolumn{10}{l}{\textit{Centered and renormalized (diagnostic contrast)}} \\
Centered & +8.87$^{\dagger}$ & +1.59 & +3.28 & +1.44 & & $-$0.77 & $-$2.63 & $-$1.88 & $-$2.68 \\
Centered, length terciles & +3.05$^{*}$ & +0.44 & +0.73 & +0.22 & & $-$0.41 & $-$1.74 & $-$1.25 & $-$2.08 \\
Centered, cross-thread & +8.78$^{\dagger}$ & $-$0.39 & +1.54 & -- & & $-$0.76 & $-$2.51 & $-$1.78 & -- \\
Centered, author-balanced & +3.72$^{\dagger}$ & +2.23 & +3.24 & -- & & $-$0.29 & $-$1.21 & $-$0.72 & -- \\
\bottomrule
\end{tabular}
\caption{Right $\times$ post difference-in-differences (right-flaired users in r/AskConservatives against left-flaired users in r/AskALiberal), in SD units of each series in 2021--22. The corrected measure averages uncentered pairwise cosines within global word-count quintiles. Newey--West standard errors with six lags. $^{**}p<.01$, $^{*}p<.05$, $^{\dagger}p<.10$. Columns other than ``mpnet (all)'' use the stratified sample of 915{,}225 comments, and ``mpnet (all)'' uses all 6{,}056{,}593.}
\label{tab:did}
\end{table*}

Figure~\ref{fig:forest} and Table~\ref{tab:did} collect the asymmetry estimate across three encoders, eight measures, and two baselines. With the full 2019 baseline, the original full-corpus estimate is 3.26 SD ($p=0.002$) and the DeBERTa sample reproduces it at 3.96 ($p=0.021$), which confirms that the sample is faithful. The length-only placebo produces 3.75 ($p=0.046$), statistically indistinguishable from the real series. The corrected measure removes the asymmetry under every encoder, to 1.09 SD under DeBERTa ($p=0.26$), 0.08 under mpnet ($p=0.95$), and 0.44 under BGE ($p=0.72$), and residualizing on the placebo gives the same picture ($-0.58$ to $+0.59$, every $p>0.4$). The centered measures behave less predictably. Centered DeBERTa without length control remains large but imprecise (8.9 SD, $p=0.06$), which is the U-shape from the truncation experiment at work, and adding length terciles after centering still leaves 3.1 ($p=0.04$).

The last column of Table~\ref{tab:did} recomputes the mpnet measures on every one of the 6.06 million comments. The corrected full-corpus estimate is 0.00 SD ($p=1.00$) with the 2019 baseline and 1.31 ($p=0.16$) with the 2021 baseline, against 0.08 and 1.31 on the sample. The raw estimates are 0.63 and 1.21 against 0.67 and 1.23, and the centered ones 1.44 and $-2.68$ against 1.59 and $-2.63$. The monthly full-corpus and sample series correlate at $r=0.975$--$0.998$ in every group and measure, so sampling drives none of these results.

When the pre-trend is estimated from 2021--22 only, after the early collapse in comment length, the original full-corpus estimate falls to 1.03 SD ($p=0.18$) and no measure under any encoder produces an asymmetry with $p<0.10$. The corrected estimates are positive and small (0.67--1.31 SD, $p\ge0.13$), while every centered sentence-encoder estimate is negative ($-1.2$ to $-2.7$). The observed asymmetry comes from the decline of 2019--20.

Under the corrected full-corpus measure with a 2021--22 baseline, all four groups drift above their pre-trend after 2022, by 2.1--3.2 SD of out-of-sample departure, and the conservative group shows the smallest drift (2.05 SD against 2.46 for left-flaired users in r/AskALiberal). The month-to-month movements are strongly synchronized within each subreddit ($r=0.74$--$0.79$ between its Left and Right series) and weakly across them ($r=0.22$--$0.46$), which points to a driver at the level of the subreddit.

\section{Diagnostic 5: Where in the Spectrum the Artifact Lives}

The pairwise mean can be taken apart exactly. Let $v_1,\dots,v_d$ be the eigenvectors of the pooled second-moment matrix of all embeddings, let $m$ be a cell's mean vector, and let $w_k$ be the cell's mean squared projection on $v_k$. Then
\begin{equation}
H=\sum_{k=1}^{d}\frac{n\,(m^\top v_k)^2-w_k}{n-1},
\end{equation}
so homogeneity is a sum of contributions from orthogonal directions, and any band of directions has its own series and its own difference-in-differences. We group the directions into the first component, components 2--5, 6--20, 21--100, and the remainder.

\begin{figure}[t]
\centering
\includegraphics[width=\columnwidth]{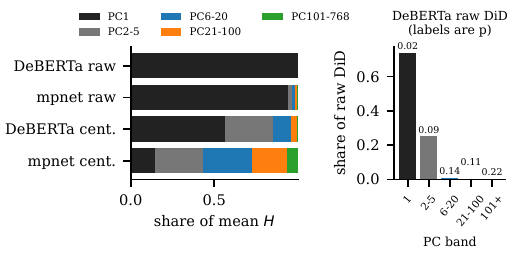}
\caption{Spectral decomposition of $H$. Left, share of mean homogeneity carried by each band of principal directions of the pooled second moment. Right, share of the raw DeBERTa Right $\times$ post difference-in-differences (2019 baseline) carried by each band, with $p$-values.}
\label{fig:spec}
\end{figure}

The first direction carries 99.9\% of mean raw homogeneity under DeBERTa (Figure~\ref{fig:spec}), 99.4\% under BGE, and 94.4\% under mpnet, although it holds 92.5\%, 52.5\%, and 12.6\% of the second moment. For DeBERTa and BGE this band also carries the length signal, since its monthly series correlates with mean log length at 0.78 and 0.88. Of the raw DeBERTa estimate of 0.0089, the first direction contributes 0.0065 ($p=0.017$) and components 2--5 another 0.0022 ($p=0.09$), and no higher band contributes anything distinguishable from zero. Under mpnet the length signal sits lower, in components 6--20 ($r=0.59$), which is why mpnet behaves differently in every diagnostic above.

After centering, the first direction still carries 56\% of DeBERTa homogeneity and components 2--5 another 29\%, and both band series correlate with length at 0.84 and 0.83, so centering moves the length signal into whichever directions lead after it. The large and imprecise centered DeBERTa estimate in Table~\ref{tab:did} comes from these two bands. Removing directions is therefore a weaker remedy than handling length. Projecting the first direction out of the raw DeBERTa embeddings, with no renormalization, leaves a noisy series whose estimate swings from 15.1 SD ($p=0.09$) with the full baseline to $-2.3$ SD ($p=0.03$) with the later one. Stratifying the raw, uncentered measure by length quintiles, by contrast, brings the DeBERTa asymmetry from 3.96 SD ($p=0.021$) to 1.09 ($p=0.26$), and the BGE and mpnet estimates to 0.44 and 0.08 ($p>0.7$), with nothing else changed.

An additive multi-scale decomposition of the daily corrected series shows that the four groups share variation mainly at scales above 512 days ($r=0.38$--$0.99$ between groups, and $0.97$--$0.99$ among the three smaller groups), while at the one-to-four-month scale of news cycles their coherence is 0.0--0.5. Homogeneity correlates with length mainly in the same slowest component ($r=0.79$--$0.98$ in three groups and 0.09 among left-flaired users in r/AskALiberal). That slow drift is precisely what an interrupted time series uses to fit its pre-trend, and so the component of the series that is most contaminated by length is the one that decides the estimate.

\section{Diagnostic 6: Shared Agenda Reads as Shared Voice}

Once length and anisotropy are handled, a centered cross-thread similarity still rises whenever many people discuss the same news, whether or not they discuss it in the same way. We separate the two on a weekly panel of 366 weeks built from the centered mpnet embeddings. For each subreddit-week, the agenda term is the mean squared projection of that week's comments onto their first principal direction. The residual similarity is computed after that direction is removed, and the within-thread similarity is computed among comments replying to the same submission.

The agenda term alone explains most of the weekly variance in cross-thread similarity, with $R^2=0.78$ in r/AskALiberal and $0.44$ in r/AskConservatives. In the 2021--25 window, the post-December-2022 level shift is not positive in either forum (AskALiberal $+0.0013$ with $p=0.41$, AskConservatives $-0.0033$ with $p=0.04$), and controlling for the agenda term shrinks both toward zero ($+0.0007$ and $-0.0013$, $p>0.16$).

\begin{figure}[t]
\centering
\includegraphics[width=\columnwidth]{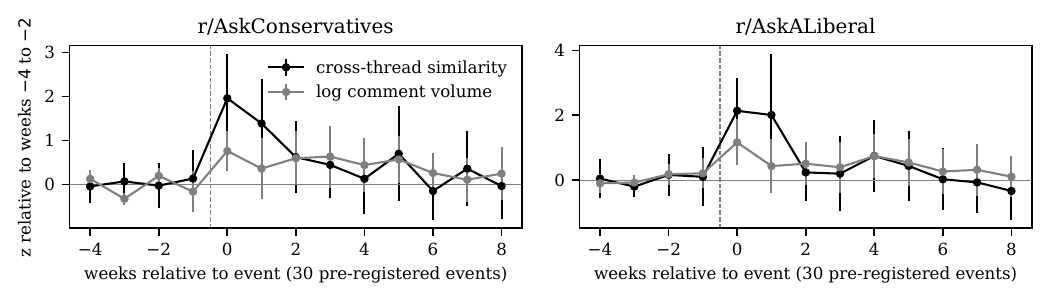}
\caption{Weekly cross-thread similarity (centered mpnet) and log comment volume around 30 pre-registered U.S. political events, with 95\% intervals.}
\label{fig:events}
\end{figure}

For an event study we fixed 30 U.S. political events in advance (Figure~\ref{fig:events}), from the first Trump impeachment inquiry to the assassination of Charlie Kirk, detrended each weekly series with a rolling 26-week median, and scaled it by its median absolute deviation. Relative to weeks $-4$ to $-2$, cross-thread similarity rises by 1.3 to 2.1 $z$ in every subreddit-by-ideology cell in the event week and the week after, and none of 500 placebo draws of random non-event weeks produces an effect that large ($p<0.002$). The effect is gone by the sixth week. The agenda term carries it ($+1.70$ $z$ in r/AskALiberal, $+1.25$ in r/AskConservatives), while residual similarity ($-0.24$, $-0.15$) and within-thread similarity ($-0.12$ in both) do not move.

The largest single shock in r/AskALiberal is the week of 7 October 2023 ($+13.3$ $z$), and the next largest fall in the summer and autumn of 2024, all inside what the original design called the generative-AI era. A post-2022 indicator in a cosine series absorbs the news cycle of those years unless the measure separates a shared topic from a shared voice. We treat this diagnostic as a demonstration of the confound and do not claim to have resolved it.

\section{Downstream Inferences Change Too}

The original study concluded that the shift vanished among authors active throughout the period and was carried by newly entering cohorts. Its stayer analysis used the DeBERTa [CLS] token at 128 tokens while the headline series used mean pooling at 256, so the two results were never measured on the same scale. On the corrected measure the stayer series shows no shift either, $-0.10$ SD among right-flaired users in r/AskConservatives with the 2019 baseline and 0.33 with the 2021 baseline ($p=0.73$ and $0.29$). With no aggregate effect left to explain, a null among stayers says nothing about mechanism. Entrants did change the composition of the forums. Authors whose first comment in a group came after December 2022 wrote 73\% of post-period comments in both r/AskConservatives groups and 42\% among left-flaired users in r/AskALiberal, and their comments are slightly more alike than stayers' comments in the same month (by 0.005--0.014, $p<0.001$) on both sides, so they cannot produce an asymmetry. A selection test on the pre-joining Reddit histories of 4{,}198 entrants, scored with the corrected measure, finds no cohort-specific shift once group trends are included ($p=0.35$).

\section{A Second Setting: Stack Exchange}

We repeat the core diagnostics on Stack Exchange, a platform with different text, different users, and a well-documented ChatGPT shock \citep{delriochanona2024}. From the April 2024 Stack Exchange dump we take every answer posted between January 2018 and March 2024 on eight sites, strip code blocks and markup, and sample up to 2{,}000 answers per site and month, which yields 575{,}000 answers. Four sites are technical, where ChatGPT substitutes directly for asking and answering (Super User, Unix \& Linux, Software Engineering, Cross Validated), and four are humanities and experience sites (English, History, Philosophy, Travel). We encode every answer with the same three encoders and build the same raw, centered, length-stratified, and length-placebo series, with placebo donors drawn within $\pm5\%$ of each answer's word count.

\begin{figure}[t]
\centering
\includegraphics[width=\columnwidth]{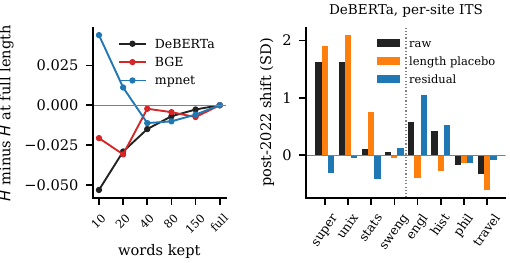}
\caption{Stack Exchange. Left, $H$ of the same 5{,}000 long answers truncated to $k$ words, relative to full length. Right, post-November-2022 interrupted-time-series shifts of raw DeBERTa $H$, its length-only placebo, and the placebo residual, in SD units of each site's 2021--22 series. Technical sites are left of the dotted line.}
\label{fig:se}
\end{figure}

Random answer pairs share a cosine of 0.93 under DeBERTa, 0.48 under BGE, and 0.02 under mpnet, and truncating long answers raises DeBERTa homogeneity from 0.902 at ten words to 0.955 at full length (Figure~\ref{fig:se}). Across the 75 months, raw DeBERTa homogeneity correlates with its placebo at a median of 0.63 across sites, 0.53 in first differences, and 0.96 and 0.81 on the two largest technical sites. Those two sites show the kind of result that would be reported as AI-era homogenization. Raw homogeneity jumps by 1.62 SD on Super User and 1.63 SD on Unix \& Linux after November 2022 ($p<0.001$). The placebo jumps by 1.90 and 2.09 SD, and the residual is $-0.32$ and $-0.05$. Answers on both sites grew longer after ChatGPT, from 103 to 109 and from 95 to 110 words on average. On Reddit the artifact came from shrinking comments, and here it comes from growing answers, which the measure registers in the same way as a change in content.

For technical against humanities sites, the raw DeBERTa difference-in-differences is $+0.73$ SD ($p=0.048$), the placebo alone gives $+1.53$ ($p<0.001$), and the residual is $-0.50$ ($p=0.20$). Mean word count itself shifts by $+1.73$ SD ($p<0.001$). After length is handled, no encoder shows homogenization of the technical sites. The corrected estimates are null or negative, and the negative ones depend on the encoder and baseline and are driven by the humanities sites, whose own composition moved (Philosophy's monthly volume doubled against 2021--22). We read this as a second null for the substantive question and a second positive case for the artifact. The direction of the artifact also differs by encoder. Under DeBERTa homogeneity rises with length, under mpnet it falls with length up to about 40 words, and BGE is nearly flat on these answers, so a study cannot predict the sign of the bias without running the diagnostics.

\section{Where the Risk Lies}

We searched arXiv, the web, and the reference lists of the seed papers for studies that measure homogenization or diversity with embedding similarity, and we coded 31 that we could locate and, in most cases, verify from the full text. For each we recorded the encoder, pooling, whether vectors were centered or whitened, whether length was handled, and whether the design compares periods. We then rated each study's exposure to the artifact documented here. The rating is our own and measures exposure. A high rating means that the conditions for the artifact are present, and only the authors' data can show whether a given result survives.

\begin{table}[t]
\centering
\small
\begin{tabular}{lcccc}
\toprule
& \multicolumn{2}{c}{Cross-sectional} & \multicolumn{2}{c}{Time series} \\
Encoder family & $n$ & at risk & $n$ & at risk \\
\midrule
MLM or decoder & 4 & 3 & 1 & 1 \\
API embedding & 5 & 0 & 4 & 2 \\
Sentence encoder & 9 & 0 & 5 & 3 \\
Other or unclear & 1 & 1 & 2 & 0 \\
\bottomrule
\end{tabular}
\caption{Census of 31 embedding-similarity studies of homogenization, by encoder family and design. At risk counts studies rated medium or higher.}
\label{tab:census}
\end{table}

Four of the five studies that use masked or decoder language models as encoders are rated medium or higher (Table~\ref{tab:census}), including the original Reddit study, a comparison of human and model-written admission essays on T5 mean-pooled embeddings \citep{lee2025essays}, a study of essay writing with decoder embeddings where text length differs across conditions \citep{kosmyna2025}, and a BERT-based study of group homogeneity in generated text \citep{lee2024homog}. Among cross-sectional studies with sentence or API encoders, none is rated at risk. The remaining exposure lies in time-series and before-after designs, where even a well-behaved encoder meets drifting text length, as in studies of model releases \citep{patel2026}, petitions \citep{corpus2025}, online reviews \citep{lerman2026}, and admissions essays across application cycles \citep{moon2025essays}. Only one of the 12 time-series designs centers its embeddings \citep{burtch2026}, and none that we could verify adjusts the similarity itself for length.

Doshi and Hauser's experiment on AI-assisted short stories \citep{doshi2024} is rated low because its conditions are randomized and its encoder was trained for similarity, which makes it a check on whether the census predicts safety correctly. We re-encoded their 293 human-written stories with the three encoders used here. Their published measure, each story's similarity to the other stories in its condition, gives a homogenization effect of 0.56 human-condition SD for writers who saw one AI idea. With raw embeddings the effect survives under every encoder, at 0.29--0.50 SD for their story-level measure and 0.58--1.03 SD for the pairwise measure of this paper. Survival under DeBERTa shows that a strong experimental design protects against a weak encoder.

\begin{figure}[t]
\centering
\includegraphics[width=\columnwidth]{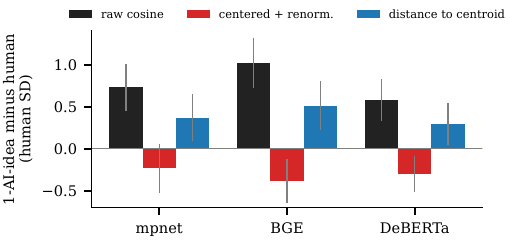}
\caption{Reverse control on the Doshi and Hauser stories. Difference in pairwise homogeneity between writers given one AI idea and the human-only condition, in SD of the human-only condition, with 95\% intervals.}
\label{fig:doshi}
\end{figure}

After centering on the pooled mean and renormalizing, the effect reverses sign under every encoder (Figure~\ref{fig:doshi}), to $-0.39$ SD under BGE ($p=0.004$), $-0.23$ under mpnet, and $-0.30$ under DeBERTa ($p=0.005$). AI-assisted stories lie closer to the pooled mean, their centered vectors are 0.31--0.61 SD shorter, and two thirds of the pooled mean comes from the AI conditions. Renormalization inflates those short residual vectors to unit length and turns closeness into noise. A measure that does not renormalize, the squared distance of each story to the leave-one-out centroid of its condition, is unchanged by centering and recovers the effect at 0.29--0.52 SD ($p\le0.022$). AI-assisted stories were also 10--17\% longer, but the effect survives under DeBERTa with length controlled (0.21 SD, $p=0.06$). Whenever the conditions being compared differ in how typical their texts are, as groups and periods on Reddit can, centering with renormalization can manufacture an effect of either sign.

\section{What the Corpus Does Show}

Within-group homogeneity did not break after ChatGPT in either camp, and we take this and the corrected results below as the substantive findings of the re-measurement. More changed between the camps. Within each subreddit, month, and length quintile we compute the squared distance between the Left and Right centroids from the same uncentered sum vectors, $D=h_{LL}+h_{RR}-2h_{LR}$, where $h_{LR}$ is the mean cosine across camps, and average over quintiles. We also estimate how well a logistic classifier separates Left from Right comments once length is projected out.

\begin{figure}[t]
\centering
\includegraphics[width=\columnwidth]{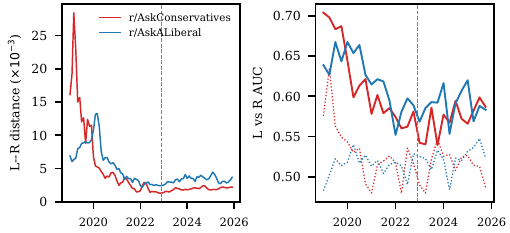}
\caption{Left, monthly squared distance between Left and Right centroids on all comments (uncentered mpnet, within length quintiles, three-month moving average). Right, quarterly AUC of a Left versus Right classifier on length-residualized embeddings (solid) and on length alone (dotted).}
\label{fig:between}
\end{figure}

In r/AskConservatives the Left--Right distance fell from $14.5\times10^{-3}$ in 2019 to $1.8\times10^{-3}$ in 2022 (Figure~\ref{fig:between}), and in r/AskALiberal from $8.0\times10^{-3}$ to $2.5\times10^{-3}$. After December 2022 the distance is flat in r/AskConservatives and drifts slightly up in r/AskALiberal, so the camps converged before ChatGPT and stopped converging before it arrived. Separability fell with the distance, from an AUC of 0.69 to 0.57 and from 0.64 to 0.58, while length alone separates the camps at 0.51. On the sample the decline appears under mpnet (a factor of 7.8 in r/AskConservatives and 3.6 in r/AskALiberal) and BGE (6.0 and 3.5), and under DeBERTa in r/AskConservatives (8.5). A length placebo that keeps each camp's length mix produces distances below $1\times10^{-3}$ in every encoder and year. We have no test of why the camps converged and treat the decline as a description of these two forums.

Who converged is harder to say. Pooling each year and excluding pairs written by the same author, the distance among users already active in 2019 fell from $4.5\times10^{-3}$ to $1.6\times10^{-3}$ by 2022 in r/AskConservatives and to about zero in r/AskALiberal, close to the fall among all users ($4.5$ to $0.8$ and $4.7$ to $1.5$). The bootstrap intervals for the early cohorts span several units of $10^{-3}$, so the data cannot apportion the convergence between incumbents and newcomers.

We also sampled 192{,}975 direct replies and compared each reply's similarity to its parent with its similarity to a control comment from the same thread, camp, and length tercile written by a third author. Replies are far closer to the comment they answer (accommodation 0.23 on uncentered mpnet), and replies across the divide accommodate more than replies within a camp. Relative to left-flaired users answering left-flaired users, right-flaired users answering left-flaired users accommodate by $+0.017$, the reverse direction by $+0.006$, and right-flaired users answering each other by $-0.006$ ($p<0.001$). Cross-camp replies quote their parents more often (19.5\% against 13.6--14.8\%), and without quoting replies the ordering survives but shrinks ($+0.009$ and $-0.008$, $p<0.001$). Accommodation did not rise after ChatGPT. Without quoting replies it fell slightly in every reply type ($-0.009$, $p=0.052$).

Splitting each cell's pairwise mean exactly into pairs from the same thread or author and unrelated pairs, within each length quintile, shows that same-thread pairs, under 0.6\% of all pairs, are about twice as similar as unrelated pairs. Of the fall of 0.0056 in corrected homogeneity among right-flaired users in r/AskConservatives between 2019--20 and 2021--22, less than a third ($-0.0016$) comes from unrelated pairs becoming less alike, and the rest from cells becoming less dominated by a few prolific authors and busy threads, so the founding-era decline that the original design read as a trend is mostly composition. After December 2022 same-thread similarity in r/AskConservatives fell relative to r/AskALiberal ($-0.030$, $p=0.010$ with a 2021 baseline), as comments per thread rose from 28.5 to 42.9. Under a test that over-rejects we read this as suggestive evidence of the opposite of homogenization.

\section{A Protocol for Embedding-Based Homogeneity}

Before reporting a trend, report the random-pair baseline, the variance share of the leading principal components, and the share of homogeneity that each band of directions carries. If unrelated comments already share a cosine of 0.9, or a single direction carries nearly all of the measure, the measure has almost no room to register meaning, and readers deserve to know that before they see the trend \citep{ethayarajh2019,rudman2022isoscore}.

Use encoders trained for sentence similarity, and treat mean-pooled outputs of masked language models as unsuitable for this purpose. Sentence-BERT was built to fix exactly this failure \citep{reimers2019sbert}.

Handle length explicitly, through stratification, reweighting, or residualization, and treat centering as a step that needs its own checks. Centering alone leaves the U-shaped dependence documented above, and sentence encoders alone leave BGE's large monotone length effect. Centering followed by renormalization is unsafe whenever the groups or periods being compared differ in how close their texts lie to the corpus mean, because it then inflates the residual vectors of the more typical group and can reverse a genuine effect, as it did in the Doshi and Hauser stories. Prefer measures that are invariant to centering and do not renormalize, such as the squared distance of each text to its group centroid or the trace of the group covariance, and compute them within length strata. When centered and uncentered measures disagree in sign, report both and investigate before choosing.

Run the length-only placebo and report it next to the main estimate. It needs no new encoding, only a resampling of embeddings already in memory, and it answers the one question a reader should ask of every embedding-similarity trend, namely how much of it length alone would produce.

Separate agenda from voice. Report how much of the similarity series a dominant-topic term explains, and test events directly, because a post-period indicator in a time series of cosines absorbs whatever the news did in those years.

Report estimates across several encoders and several baselines, in line with long-standing advice to validate text measures against what they claim to capture \citep{grimmer2013,rodriguez2022,jacobs2021measurement}. Here the result that looked robust across designs, aggregation levels, and matching schemes was fragile across measures, because every design shared the same instrument. Finally, calibrate the test itself. Running the analysis on fixed-content synthetic series with no effect shows how often the chosen time-series design rejects by chance, which in this case was two to three times the nominal rate.

\section{Discussion}

The original analysis replicated across interrupted time series, regression discontinuity in time, trend-corrected difference-in-differences, propensity-score matching, and three temporal aggregations, which is careful by the standards of its genre. Every design took the same series as input, and the series measured length. Agreement among such designs is evidence about the estimator and says little about the construct.

We argue that the failure generalizes well beyond the study examined here, because platform text is the setting in which length shifts for reasons unrelated to the substantive question, including moderation rules, interface changes, the life cycle of a new community, and, plausibly, the arrival of AI writing tools themselves. A study that finds AI-era homogenization with a length-sensitive measure may be detecting that people write shorter or longer comments, which is a finding about style volume and says nothing yet about the range of ideas. Lexical and distributional approaches to AI exposure \citep{kobak2025,liang2024monitoring,liang2025widespread} and standardized diversity batteries \citep{shaib2025diversity} offer complementary checks that do not share this vulnerability.

We examine only two platforms, and the magnitude and even the sign of the artifact will vary with the encoder and the length distribution of other data. The census is a structured search and does not exhaust the literature, its risk ratings are our own judgment from the reported methods, and four ratings rest on incomplete information. The substantive findings rest on corrected measures that we chose after seeing the artifact, and the Newey--West test over-rejects, so we report them as descriptions of this corpus and apply no correction for the many estimates in the paper. Our corrections are simple, and whitening, cluster-based isotropy methods \citep{rajaee2021cluster}, or length-normalized encoders may do better. The agenda--voice split uses a single principal direction, and the 30 events include overlapping weeks (the 2020 election and its call, and the clustered events of mid-2024). Most re-measurement runs on a stratified sample of 915{,}225 comments, but DeBERTa on that sample reproduces the original full-corpus estimate, and mpnet on all 6.06 million comments reproduces the sample estimates for every measure and baseline. On Reddit a measure designed to detect converging discourse detected shrinking comments, and on Stack Exchange it detected growing answers.

\section{Ethical Considerations}

The corpus consists of public Reddit comments, and the entrant histories were retrieved from the public Arctic Shift archive. The Stack Exchange answers come from the public data dump released under CC BY-SA, and the Doshi and Hauser stories from their public replication archive on Dryad. The analysis uses only aggregate statistics at the level of subreddit, ideology flair, and month or week. We do not quote, identify, or profile individual users, and the entrant histories enter only as averaged embeddings. Flair is self-assigned and serves here only as a grouping variable, with no claim about any individual's beliefs. The analysis shows that a published claim of ideologically asymmetric homogenization, which could be used to characterize one political group as more susceptible to AI influence, does not survive re-measurement. We see no new risks to the people whose comments are analyzed beyond those of the original study.

\section*{Acknowledgments}
Claude (Anthropic) was used to write and debug analysis code and to help edit the text. The authors designed the study, checked every reported estimate against the analysis output, and is responsible for the interpretation and conclusions.

\ifanon\else
The Stack Exchange answers are used under CC BY-SA 4.0.
\fi

\bibliography{refs}

\begin{thebibliography}{41}
\providecommand{\natexlab}[1]{#1}

\bibitem[{Agarwal, Naaman, and Vashistha(2025)}]{agarwal2025}
Agarwal, D.; Naaman, M.; and Vashistha, A. 2025.
\newblock {AI} Suggestions Homogenize Writing Toward {W}estern Styles and
  Diminish Cultural Nuances.
\newblock In \emph{Proceedings of CHI 2025}, 1--21.

\bibitem[{Anderson, Shah, and Kreminski(2024)}]{anderson2024}
Anderson, B.~R.; Shah, J.~H.; and Kreminski, M. 2024.
\newblock Homogenization Effects of Large Language Models on Human Creative
  Ideation.
\newblock In \emph{Proceedings of the ACM Conference on Creativity \&
  Cognition}, 413--425.

\bibitem[{Burtch(2026)}]{burtch2026}
Burtch, G. 2026.
\newblock The Hitchhiker's Guide to Monoculture: {AI} Homogenizes Syntax, Not
  (Necessarily) Semantics.
\newblock arXiv:2607.13077.

\bibitem[{Corpus et~al.(2025)Corpus, Gilbert, Koenecke, and
  Naaman}]{corpus2025}
Corpus, I.; Gilbert, E.; Koenecke, A.; and Naaman, M. 2025.
\newblock Introducing {AI} to an Online Petition Platform Changed Outputs but
  not Outcomes.
\newblock arXiv:2511.13949.

\bibitem[{del Rio-Chanona, Laurentsyeva, and Wachs(2024)}]{delriochanona2024}
del Rio-Chanona, R.~M.; Laurentsyeva, N.; and Wachs, J. 2024.
\newblock Large language models reduce public knowledge sharing on online
  {Q\&A} platforms.
\newblock \emph{PNAS Nexus}, 3(9): pgae400.

\bibitem[{Doshi and Hauser(2024)}]{doshi2024}
Doshi, A.~R.; and Hauser, O.~P. 2024.
\newblock Generative {AI} enhances individual creativity but reduces the
  collective diversity of novel content.
\newblock \emph{Science Advances}, 10(28): eadn5290.

\bibitem[{Ethayarajh(2019)}]{ethayarajh2019}
Ethayarajh, K. 2019.
\newblock How Contextual are Contextualized Word Representations? Comparing the
  Geometry of {BERT}, {ELMo}, and {GPT-2} Embeddings.
\newblock In \emph{Proceedings of EMNLP-IJCNLP 2019}, 55--65.

\bibitem[{Friedman and Dieng(2023)}]{friedman2023vendi}
Friedman, D.; and Dieng, A.~B. 2023.
\newblock The {Vendi} Score: A Diversity Evaluation Metric for Machine
  Learning.
\newblock \emph{Transactions on Machine Learning Research}.

\bibitem[{Gao et~al.(2019)Gao, He, Tan, Qin, Wang, and Liu}]{gao2019}
Gao, J.; He, D.; Tan, X.; Qin, T.; Wang, L.; and Liu, T.-Y. 2019.
\newblock Representation Degeneration Problem in Training Natural Language
  Generation Models.
\newblock In \emph{International Conference on Learning Representations}.

\bibitem[{Grimmer and Stewart(2013)}]{grimmer2013}
Grimmer, J.; and Stewart, B.~M. 2013.
\newblock Text as Data: The Promise and Pitfalls of Automatic Content Analysis
  Methods for Political Texts.
\newblock \emph{Political Analysis}, 21(3): 267--297.

\bibitem[{Hausman and Rapson(2018)}]{hausman2018rdit}
Hausman, C.; and Rapson, D.~S. 2018.
\newblock Regression Discontinuity in Time: Considerations for Empirical
  Applications.
\newblock \emph{Annual Review of Resource Economics}, 10: 533--552.

\bibitem[{He, Gao, and Chen(2023)}]{he2023debertav3}
He, P.; Gao, J.; and Chen, W. 2023.
\newblock {DeBERTaV3}: Improving {DeBERTa} using {ELECTRA}-Style Pre-Training
  with Gradient-Disentangled Embedding Sharing.
\newblock In \emph{International Conference on Learning Representations}.

\bibitem[{Jacobs and Wallach(2021)}]{jacobs2021measurement}
Jacobs, A.~Z.; and Wallach, H. 2021.
\newblock Measurement and Fairness.
\newblock In \emph{Proceedings of ACM FAccT 2021}, 375--385.

\bibitem[{Kobak et~al.(2025)Kobak, Gonz{\'a}lez-M{\'a}rquez, Horv{\'a}t, and
  Lause}]{kobak2025}
Kobak, D.; Gonz{\'a}lez-M{\'a}rquez, R.; Horv{\'a}t, E.-{\'A}.; and Lause, J.
  2025.
\newblock Delving into {LLM}-assisted writing in biomedical publications
  through excess vocabulary.
\newblock \emph{Science Advances}, 11(27): eadt3813.

\bibitem[{Kosmyna et~al.(2025)Kosmyna, Hauptmann, Yuan, Situ, Liao,
  Beresnitzky, Braunstein, and Maes}]{kosmyna2025}
Kosmyna, N.; Hauptmann, E.; Yuan, Y.~T.; Situ, J.; Liao, X.-H.; Beresnitzky,
  A.~V.; Braunstein, I.; and Maes, P. 2025.
\newblock Your Brain on {ChatGPT}: Accumulation of Cognitive Debt when Using an
  {AI} Assistant for Essay Writing Task.
\newblock arXiv:2506.08872.

\bibitem[{Lee et~al.(2025)Lee, Alvero, Joachims, and Kizilcec}]{lee2025essays}
Lee, J.; Alvero, A.; Joachims, T.; and Kizilcec, R. 2025.
\newblock Poor Alignment and Steerability of Large Language Models: Evidence
  from College Admission Essays.
\newblock arXiv:2503.20062.

\bibitem[{Lee, Montgomery, and Lai(2024)}]{lee2024homog}
Lee, M. H.~J.; Montgomery, J.~M.; and Lai, C.~K. 2024.
\newblock Large Language Models Portray Socially Subordinate Groups as More
  Homogeneous, Consistent with a Bias Observed in Humans.
\newblock In \emph{Proceedings of the 2024 ACM Conference on Fairness,
  Accountability, and Transparency (FAccT)}.
\newblock ArXiv:2401.08495.

\bibitem[{Lerman, Rigbi, and Dover(2026)}]{lerman2026}
Lerman, V.; Rigbi, O.; and Dover, Y. 2026.
\newblock The Role of {AI} in Online Reviews.
\newblock arXiv:2609.22198.

\bibitem[{Li et~al.(2020)Li, Zhou, He, Wang, Yang, and Li}]{li2020bertflow}
Li, B.; Zhou, H.; He, J.; Wang, M.; Yang, Y.; and Li, L. 2020.
\newblock On the Sentence Embeddings from Pre-trained Language Models.
\newblock In \emph{Proceedings of EMNLP 2020}, 9119--9130.

\bibitem[{Liang et~al.(2024)Liang, Izzo, Zhang, Lepp, Cao, Zhao, Chen, Ye, Liu,
  Huang, McFarland, and Zou}]{liang2024monitoring}
Liang, W.; Izzo, Z.; Zhang, Y.; Lepp, H.; Cao, H.; Zhao, X.; Chen, L.; Ye, H.;
  Liu, S.; Huang, Z.; McFarland, D.~A.; and Zou, J.~Y. 2024.
\newblock Monitoring {AI}-Modified Content at Scale: A Case Study on the Impact
  of {ChatGPT} on {AI} Conference Peer Reviews.
\newblock In \emph{Proceedings of ICML 2024}, 29575--29620.

\bibitem[{Liang et~al.(2025)Liang, Zhang, Codreanu, Wang, Cao, and
  Zou}]{liang2025widespread}
Liang, W.; Zhang, Y.; Codreanu, M.; Wang, J.; Cao, H.; and Zou, J. 2025.
\newblock The widespread adoption of large language model-assisted writing
  across society.
\newblock \emph{Patterns}, 6(12): 101366.

\bibitem[{Liu(2026)}]{liu2026asym}
Liu, F. 2026.
\newblock Asymmetric Discourse Homogenization and Shared Language Technology:
  Evidence from {R}eddit.
\newblock arXiv:2608.13674.

\bibitem[{Lopez~Bernal, Cummins, and Gasparrini(2017)}]{bernal2017its}
Lopez~Bernal, J.; Cummins, S.; and Gasparrini, A. 2017.
\newblock Interrupted time series regression for the evaluation of public
  health interventions: a tutorial.
\newblock \emph{International Journal of Epidemiology}, 46(1): 348--355.

\bibitem[{Moon et~al.(2025)Moon, Kushlev, Bank, Viskontas, and
  Green}]{moon2025essays}
Moon, K.; Kushlev, K.; Bank, A.; Viskontas, I.; and Green, A.~E. 2025.
\newblock {LLM}-Era College Admissions Essays Exhibit Paradoxical Semantic
  Trends.
\newblock OSF Preprints.

\bibitem[{Mu, Bhat, and Viswanath(2018)}]{mu2018allbutthetop}
Mu, J.; Bhat, S.; and Viswanath, P. 2018.
\newblock All-but-the-Top: Simple and Effective Postprocessing for Word
  Representations.
\newblock In \emph{International Conference on Learning Representations}.

\bibitem[{Padmakumar and He(2024)}]{padmakumar2024}
Padmakumar, V.; and He, H. 2024.
\newblock Does Writing with Language Models Reduce Content Diversity?
\newblock In \emph{International Conference on Learning Representations}.

\bibitem[{Patel et~al.(2026)Patel, Crossman, Aggarwal, and Wenger}]{patel2026}
Patel, N.; Crossman, J.; Aggarwal, E.; and Wenger, E. 2026.
\newblock Are {LLMs} Becoming Similarly Creative? Evidence from Three Years of
  Models.
\newblock arXiv:2608.19437.

\bibitem[{Rajaee and Pilehvar(2021)}]{rajaee2021cluster}
Rajaee, S.; and Pilehvar, M.~T. 2021.
\newblock A Cluster-based Approach for Improving Isotropy in Contextual
  Embedding Space.
\newblock In \emph{Proceedings of ACL-IJCNLP 2021 (Volume 2: Short Papers)},
  575--584.

\bibitem[{Reimers and Gurevych(2019)}]{reimers2019sbert}
Reimers, N.; and Gurevych, I. 2019.
\newblock Sentence-{BERT}: Sentence Embeddings using {S}iamese {BERT}-Networks.
\newblock In \emph{Proceedings of EMNLP-IJCNLP 2019}, 3980--3990.

\bibitem[{Rodriguez and Spirling(2022)}]{rodriguez2022}
Rodriguez, P.~L.; and Spirling, A. 2022.
\newblock Word Embeddings: What Works, What Doesn't, and How to Tell the
  Difference for Applied Research.
\newblock \emph{The Journal of Politics}, 84(1): 101--115.

\bibitem[{Rudman et~al.(2022)Rudman, Gillman, Rayne, and
  Eickhoff}]{rudman2022isoscore}
Rudman, W.; Gillman, N.; Rayne, T.; and Eickhoff, C. 2022.
\newblock {IsoScore}: Measuring the Uniformity of Embedding Space Utilization.
\newblock In \emph{Findings of the Association for Computational Linguistics:
  ACL 2022}, 3325--3339.

\bibitem[{Shaib et~al.(2025)Shaib, Govindarajan, Barrow, Sun, Siu, Wallace, and
  Nenkova}]{shaib2025diversity}
Shaib, C.; Govindarajan, V.~S.; Barrow, J.; Sun, J.; Siu, A.~F.; Wallace,
  B.~C.; and Nenkova, A. 2025.
\newblock Standardizing the Measurement of Text Diversity: A Tool and
  Comparative Analysis.
\newblock In \emph{Proceedings of IJCNLP-AACL 2025: System Demonstrations},
  36--46.

\bibitem[{Song et~al.(2020)Song, Tan, Qin, Lu, and Liu}]{song2020mpnet}
Song, K.; Tan, X.; Qin, T.; Lu, J.; and Liu, T.-Y. 2020.
\newblock {MPNet}: Masked and Permuted Pre-training for Language Understanding.
\newblock In \emph{Advances in Neural Information Processing Systems 33}.

\bibitem[{Sourati, Ziabari, and Dehghani(2025)}]{sourati2025}
Sourati, Z.; Ziabari, A.~S.; and Dehghani, M. 2025.
\newblock The Homogenizing Effect of Large Language Models on Human Expression
  and Thought.
\newblock \emph{Trends in Cognitive Sciences}.

\bibitem[{Steck, Ekanadham, and Kallus(2024)}]{steck2024cosine}
Steck, H.; Ekanadham, C.; and Kallus, N. 2024.
\newblock Is Cosine-Similarity of Embeddings Really About Similarity?
\newblock In \emph{Companion Proceedings of the ACM Web Conference 2024},
  887--890.

\bibitem[{Su et~al.(2021)Su, Cao, Liu, and Ou}]{su2021whitening}
Su, J.; Cao, J.; Liu, W.; and Ou, Y. 2021.
\newblock Whitening Sentence Representations for Better Semantics and Faster
  Retrieval.
\newblock arXiv:2103.15316.

\bibitem[{Timkey and van Schijndel(2021)}]{timkey2021}
Timkey, W.; and van Schijndel, M. 2021.
\newblock All Bark and No Bite: Rogue Dimensions in Transformer Language Models
  Obscure Representational Quality.
\newblock In \emph{Proceedings of EMNLP 2021}, 4527--4546.

\bibitem[{Xiao et~al.(2023)Xiao, Li, Hudson, Lin, and
  Al~Moubayed}]{xiao2023length}
Xiao, C.; Li, Y.; Hudson, G.~T.; Lin, C.; and Al~Moubayed, N. 2023.
\newblock Length is a Curse and a Blessing for Document-level Semantics.
\newblock In \emph{Proceedings of EMNLP 2023}, 1385--1396.

\bibitem[{Xiao et~al.(2024)Xiao, Liu, Zhang, Muennighoff, Lian, and
  Nie}]{xiao2024cpack}
Xiao, S.; Liu, Z.; Zhang, P.; Muennighoff, N.; Lian, D.; and Nie, J.-Y. 2024.
\newblock {C-Pack}: Packed Resources For General {C}hinese Embeddings.
\newblock In \emph{Proceedings of SIGIR 2024}, 641--649.

\bibitem[{Zhou et~al.(2022)Zhou, Ethayarajh, Card, and
  Jurafsky}]{zhou2022cosine}
Zhou, K.; Ethayarajh, K.; Card, D.; and Jurafsky, D. 2022.
\newblock Problems with Cosine as a Measure of Embedding Similarity for High
  Frequency Words.
\newblock In \emph{Proceedings of ACL 2022 (Volume 2: Short Papers)}, 401--423.

\bibitem[{Zhou et~al.(2025)Zhou, Dai, Cao, Zhang, and
  Xu}]{zhou2025lengthcollapse}
Zhou, Y.; Dai, S.; Cao, Z.; Zhang, X.; and Xu, J. 2025.
\newblock Length-Induced Embedding Collapse in {PLM}-based Models.
\newblock In \emph{Proceedings of ACL 2025 (Volume 1: Long Papers)},
  28767--28791.

\end{thebibliography}

\end{document}